\documentclass[11pt]{article}

\usepackage[preprint]{acl}
\usepackage{times}
\usepackage{latexsym}
\usepackage[T2A,T1]{fontenc}
\usepackage[utf8]{inputenc}

\newcommand{\kazakhY}{%
  {\fontencoding{T2A}\fontfamily{cmr}\selectfont\cyrery}%
}
\newcommand{\kazakhBir}{%
  {\fontencoding{T2A}\fontfamily{cmr}\selectfont
   \cyrb\cyrii\cyrr}%
}
\usepackage{inconsolata}
\usepackage{amsmath}
\usepackage{booktabs}
\usepackage{tabularx}
\usepackage{array}
\usepackage{tipa}
\usepackage{microtype}
\usepackage{tikz}
\usepackage{graphicx}
\usetikzlibrary{arrows.meta,positioning}
\usepackage{amsthm}
\theoremstyle{plain}
\newtheorem{takeaway}{Take-away}

\usepackage[english,bidi=default]{babel} % English as the main language.
\title{Universal or Language-Family-Specific Script Unification for Cross-Lingual Transfer? A Case Study on Turkic Languages}

\author{Zijie Zhang \\
  The Chinese University of Hong Kong, Shenzhen \\
  \texttt{zijiezhang@link.cuhk.edu.hk}}

\begin{document}

\maketitle
\begin{abstract}
Closely related languages written in different scripts expose little surface overlap to multilingual models, limiting cross-lingual transfer. We compare two approaches to script unification: the general-purpose uroman romanizer and the family-specific Common Turkic Script (CTS). We train matched fastText models on transliterated Wikipedia corpora from 11 Turkic languages and evaluate them on WikiANN named entity recognition and Universal Dependencies part-of-speech tagging. CTS and uroman show no significant difference on NER, while both substantially outperform the official monolingual fastText baselines. POS results reveal no universal winner: language-specific differences are associated with the cross-lingual character $n$-gram coverage induced by each representation, while within-language coverage becomes more important when target-language supervision is available. Although CANINE-c achieves higher overall POS averages, the substantially simpler fastText-based systems remain competitive on several treebanks. Overall, the effectiveness of script unification depends on the language, the induced subword overlap, and the available supervision.
\end{abstract}

\section{Introduction}

Multilingual models benefit from shared lexical and subword structure across languages. Yet closely related languages may expose little of this structure when they are written in different scripts. Turkish \textit{bir} and Kazakh \textit{\kazakhBir}, for example, are cognate forms meaning ``one,'' but their written forms share no Unicode characters. This script barrier can prevent models from exploiting relationships that are linguistically present but computationally hidden.  

A shared writing system can reduce this barrier, but its design involves a fundamental trade-off. A universal romanizer may maximize broad character sharing by mapping many scripts into a small common inventory. A family-specific system may instead preserve contrasts and correspondences that are important within a particular language family. Greater character sharing and greater linguistic faithfulness are therefore not necessarily the same objective.  

Previous work has shown that common-script representations can improve multilingual transfer, especially for related and lower-resource languages \citep{nguyen-chiang-2017-transfer,khatri2020study, sun-etal-2022-alternative,moosa-etal-2023-transliteration, jayakumar-etal-2026-scripts}. However, it remains unclear whether universal or family-specific unification is preferable, and which properties of the resulting representation determine downstream transfer. We therefore ask: when closely related languages are separated by
heterogeneous scripts, which form of script unification---universal
romanization or family-specific unification---supports more effective
cross-lingual transfer, and what properties of the resulting
representations explain their performance?

We study this question in 11 Turkic languages written across Latin, Cyrillic, and Perso-Arabic scripts. We compare the \textbf{general-purpose \texttt{uroman}} romanizer \citep{hermjakob-etal-2018-box} with the \textbf{family-specific Common Turkic Script} (CTS) \citep{internationalturkicacademy2024common,
hakimov2026turkicnlpnlptoolkit}. Starting from matched corpora and task data, we create two representational views that differ primarily in transliteration, train character-$n$-gram-based fastText models \citep{bojanowski2017enriching}, and evaluate them on named entity recognition and part-of-speech tagging. We then measure the cross-lingual and within-language character $n$-gram overlap induced by each representation.  

Our results~\footnote{To support reproducibility, at
\url{https://github.com/anonymous-res-user/Turkic-Unified-Writing-System-fastText}, we release the CTS- and uroman-transliterated Wikipedia corpora, trained fastText models, training configurations, and all code scripts. The repository is now under construction.}
yield three main findings. First, CTS and uroman show no significant
difference on NER, although both outperform language-specific systems built
from officially released monolingual fastText embeddings
\citep{grave2018learning}. Second, neither representation universally
dominates POS tagging: uroman performs better in the Azerbaijani and Tatar
zero-shot settings, whereas CTS transfers more effectively to Uyghur.
Third, these differences align with the task-relevant character $n$-gram
coverage induced by each representation, while target-language supervision
shifts the balance toward within-language overlap. Although CANINE-c
achieves stronger overall POS averages, the substantially simpler
fastText-based systems remain competitive on several treebanks.

\section{Preliminaries}

\subsection{Problem Definition}

% \paragraph{.}
\paragraph{Script Unification.}
Let $\mathcal{L}$ denote a set of related languages, where each
language $\ell \in \mathcal{L}$ may use a different writing
system with character inventory $\Sigma_{\ell}$. We define
\emph{script unification} as a set of mappings
$f_{\ell}: \Sigma_{\ell}^{*} \rightarrow \Sigma_{U}^{*}$ that
convert these heterogeneous writing systems into a shared
representation $\Sigma_{U}$. The goal is not to translate the
text or erase language-specific differences, but to reduce
orthographic fragmentation so that related words and morphemes
expose comparable character and subword patterns.

\paragraph{Cross-Lingual Transfer through Script Unification.}
Script unification provides a representational bridge for
cross-lingual transfer. By making shared lexical, morphological,
and character-level patterns visible across writing systems, it
allows representations and task supervision learned from source
languages to benefit a target language. This includes zero-shot
transfer, where no target-language task supervision is available,
and multilingual joint learning, where supervision from other
languages complements the available target-language data.

% However, greater surface overlap is not necessarily more useful.
% An aggressive mapping may collapse linguistically meaningful
% distinctions or create spurious correspondences between unrelated
% forms. The relevant question is therefore not only how much
% overlap a unified representation induces, but whether that overlap
% supports downstream cross-lingual transfer.

\begin{table*}[t]
\centering
\small
\setlength{\tabcolsep}{5pt}
\renewcommand{\arraystretch}{0.8}
\begin{tabularx}{\textwidth}{
    >{\raggedright\arraybackslash}p{3.0cm}
    >{\raggedright\arraybackslash}p{3.2cm}
    >{\raggedright\arraybackslash}p{3.2cm}
    >{\raggedright\arraybackslash}X
}
\toprule
\textbf{Original Contrast or Correspondence}
&
\textbf{Universal Romanization: uroman}
&
\textbf{Family-Specific Unification: CTS}
&
\textbf{Interpretation}
\\
\midrule

Turkish \textit{\i{}} and \textit{i}
&
Both characters are mapped to \textit{i}.
&
The distinction between \textit{\i{}} and \textit{i} is preserved.
&
The uroman transliteration reduces the character inventory and increases
surface-form sharing, but merges a linguistically meaningful
contrast. CTS preserves this Turkic-specific distinction.
\\
\midrule

Kazakh \kazakhY{} and Turkish \textit{\i{}}
&
Kazakh \kazakhY{} is mapped to \textit{y}, whereas Turkish
\textit{ı} is mapped to \textit{i}.
&
Kazakh \kazakhY{} is mapped to \textit{\i{}}, matching Turkish
\textit{\i{}}.
&
In many inherited cognates, the vowels represented by Kazakh
\kazakhY{} and Turkish \textit{\i{}} reflect the same Proto-Turkic
vowel. The uroman transliteration represents them differently,
whereas CTS maps both to \textit{\i{}}, encoding a recurrent
family-internal correspondence.
\\
\midrule

Azerbaijani \textit{\textschwa} and \textit{e}
&
Both characters are mapped to \textit{e}.
&
Azerbaijani \textit{\textschwa} is represented as \textit{ä}, whereas
\textit{e} remains \textit{e}.
&
The uroman transliteration creates a more compact representation by merging the two vowels. CTS preserves the contrast between them, reducing representational collisions within Azerbaijani.
\\
\bottomrule
\end{tabularx}

\caption{
Concrete examples of universal and family-specific script
unification. The uroman transliteration prioritizes broad applicability and
a compact shared inventory, whereas CTS uses
Turkic-specific correspondences to align related sounds while
preserving linguistically relevant distinctions.
}
\label{tab:uroman-cts-examples}
\end{table*}

% These examples illustrate that the two paradigms induce different
% forms of representational sharing. Universal romanization seeks
% broader surface overlap across heterogeneous writing systems,
% whereas family-specific unification seeks more linguistically
% faithful alignment within a particular language family. The
% central question in this work is therefore not simply which
% method produces more character overlap, but which type of overlap
% provides more useful subword sharing for downstream cross-lingual
% transfer.

% The two paradigms differ not in whether they enable
% cross-lingual transfer, but in the type of sharing they induce.
% Universal romanization prioritizes broad surface overlap across
% writing systems, whereas family-specific unification prioritizes
% linguistically informed alignment within a language family.
% The central question is therefore whether downstream transfer
% benefits more from broader sharing or from more faithful
% family-level correspondences.

\subsection{Two Paradigms of Script Unification}
Approaches to script unification can be broadly divided into two
paradigms. \emph{Universal script normalization} applies a
general-purpose, language-independent mapping to place many
writing systems into a compact shared inventory. It prioritizes
broad applicability and surface-form sharing. In contrast,
\emph{family-specific unification} uses correspondences designed
for a particular language family, seeking to align related forms
while preserving distinctions that are important within that
family.
% Both paradigms aim to facilitate cross-lingual transfer, but they
% induce different forms of representational sharing. Their
% comparison therefore tests whether downstream transfer benefits
% more from broad universal overlap or from linguistically informed
% family-level alignment, and how this trade-off changes across languages and supervision regimes.

\paragraph{Type I: Universal Romanization.}
The general-purpose uroman romanizer instantiates universal script
normalization by converting diverse writing systems into a shared
Latin-script representation \citep{hermjakob-etal-2018-box}. Because it is language-independent and general-purpose, uroman can provide broad coverage using a compact inventory of basic Latin letters. However, it does not explicitly model correspondences within a particular language family. Consequently, historically or phonologically related sounds may remain differently represented, while distinct sounds may be mapped to the same form.

\paragraph{Type II: Family-Specific Unification.}
The Common Turkic Alphabet (CTA) is a 34-letter Latin-based framework designed specifically for the Turkic language family \citep{internationalturkicacademy2024common}. \texttt{TurkicNLP} operationalizes CTA as an automatic transliteration target called the Common Turkic Script (CTS) \citep{hakimov2026turkicnlpnlptoolkit}. Unlike uroman, CTS uses Turkic-specific correspondences to align related sounds across languages while preserving contrasts that may be important within individual languages.

Table~\ref{tab:uroman-cts-examples} illustrates representative
differences between the two strategies. We next compare them under
controlled experimental conditions.

\section{Methodology}

\subsection{Experimental Design}

\paragraph{Controlled Comparison.}
Our goal is to determine which script-unification strategy creates more
useful character-level sharing for joint training across related
languages. Starting from the same Wikipedia corpora for 11 Turkic
languages, we construct two multilingual training corpora: one obtained
with the general-purpose uroman romanizer and the other with the
family-specific Common Turkic Script (CTS). The two conditions use the
same languages, source snapshot, shared preprocessing, and downstream
evaluation procedure. The intended experimental factor is therefore the
strategy used to unify the writing systems.

\paragraph{Character-Level Representation Learning.}
We use fastText because it constructs word representations from character
$n$-grams and shares parameters across words with overlapping character
sequences \citep{bojanowski2017enriching}. This makes it well suited for
testing whether the character-level overlap induced by CTS or uroman
supports cross-lingual transfer.

\paragraph{Character-Sharing Analysis.}
We evaluate the two learned representations on multilingual NER and POS
tagging. These downstream tasks measure whether the character-level
sharing induced during fastText training translates into practical transfer
performance. To interpret language-specific differences, we additionally
compare the cross-lingual and within-language character $n$-gram coverage
created by the two representations. Cross-lingual coverage measures the
overlap between a target language and the other training languages, whereas
within-language coverage measures train--test overlap inside the target
language.

\subsection{Implementation}
\paragraph{Corpus Construction.}
\label{sec:script-unification}

We constructed parallel CTS and uroman corpora for 11 Turkic-language Wikipedia editions from the same January~1, 2026 Wikimedia dump snapshot: Azerbaijani (\texttt{az}), Bashkir (\texttt{ba}), Chuvash (\texttt{cv}), Karakalpak (\texttt{kaa}), Kazakh (\texttt{kk}), Kyrgyz (\texttt{ky}), Tatar (\texttt{tt}), Turkmen (\texttt{tk}), Turkish (\texttt{tr}), Uyghur (\texttt{ug}), and Uzbek (\texttt{uz}). Both conditions used identical preprocessing and postprocessing. Before transliteration, the data were normalized to NFC \citep{unicode-uax15}, stripped of trailing carriage-return and newline characters, and canonicalized by mapping apostrophe-like characters to the ASCII apostrophe. After transliteration, the data were again normalized to NFC, tokenized
according to the Unicode word-boundary rules in UAX \#29
\citep{unicode-uax29}, and serialized with single spaces. We then applied
identical line concatenation, punctuation augmentation, shuffling, and
merging procedures to both conditions. The two conditions therefore differed only in the representation-specific transliteration step.

For the uroman condition, no language code was supplied as the parameter. For the CTS condition, \texttt{Script.COMMON\_TURKIC} was always set as the target script \citep{hakimov2026turkicnlpnlptoolkit}. Because Wikipedia editions may contain material written in multiple scripts, CTS processing sequentially covered Latin, Cyrillic, and Perso-Arabic text. We used source-language-specific transliterators where available and coverage-oriented proxy mappings otherwise. See Appendix~\ref{app:data-processing} for  details.

\paragraph{Training of fastText.}

We trained matched fastText models, \texttt{cts\_ft} and \texttt{uroman\_ft}, on the corresponding final merged CTS and uroman Wikipedia corpora. With the minimum word count set to 5, the retained vocabularies yielded
2,691,885 distinct character $n$-gram types for $n=2,\ldots,5$ in the CTS
corpus and 1,913,731 in the uroman corpus.

In fastText, character $n$-grams are hashed into a fixed number of buckets
\citep{bojanowski2017enriching}. Because the two corpora contained different numbers of distinct $n$-gram types, we matched the ratio of bucket capacity to the observed $n$-gram inventory rather than imposing the same absolute bucket size. The CTS model used 21,000,000 buckets, giving a bucket-to-type ratio of 7.801, whereas the uroman model used 15,000,000 buckets, giving a ratio of 7.838. These nearly equal ratios maintained comparable relative bucket capacity and hashing pressure across the two representation conditions.

Both models used the standard, position-independent CBOW implementation provided by the public fastText library, rather than the position-weighted CBOW variant used to train the released fastText vectors for 157 languages \citep{grave2018learning}. Apart from the bucket size, the two models used identical hyperparameters. The complete training configurations are reported in Appendix~\ref{app:fasttext-configuration}.

\paragraph{Downstream Evaluation.}

We evaluated the learned representations on named entity recognition
(NER) and part-of-speech (POS) tagging. For NER, we used WikiANN data
\citep{pan-etal-2017-cross,rahimi-etal-2019-massively} for 10 of the
11 languages included in our fastText training corpora. Karakalpak (\texttt{kaa}) was the only fastText training
language not represented in the NER evaluation. For POS tagging, we used
19 UD v2.18 treebanks \citep{nivre-etal-2020-universal} spanning seven
languages: Azerbaijani (\texttt{az}), Kazakh (\texttt{kk}), Kyrgyz
(\texttt{ky}), Turkish (\texttt{tr}), Tatar (\texttt{tt}), Uyghur
(\texttt{ug}), and Uzbek (\texttt{uz}). Sentence counts for the POS data
splits are reported in Appendix~\ref{app:pos-data-splits}.

We applied the script-unification and normalization pipeline described for corpus construction to every training, validation, and test split, producing four matched datasets: \texttt{cts-ner}, \texttt{uroman-ner}, \texttt{cts-pos}, and \texttt{uroman-pos}. For each task and representation condition, all available language- or treebank-specific training splits were pooled into a single multilingual training set, and all available validation splits were likewise pooled into a single multilingual validation set. No language or treebank identity was provided to the models as an input feature. At test time, the splits remained separate: NER was evaluated independently for each WikiANN language test split, and POS was evaluated independently for each UD treebank test split.

For NER, we trained matched BiLSTM--CRF taggers \citep{lample-etal-2016-neural} using \texttt{cts\_ft} and \texttt{uroman\_ft} embeddings for the corresponding CTS and uroman conditions. For POS tagging, we analogously trained matched BiLSTM taggers. Within each task, both CTS and uroman used identical architectures and hyperparameters; only the input representation and corresponding fastText embeddings differed. For each of the \texttt{cts-ner}, \texttt{uroman-ner}, \texttt{cts-pos}, and \texttt{uroman-pos} training, we ran in five seeds: \texttt{1}, \texttt{41}, \texttt{42}, \texttt{43}, and \texttt{72}.

\subsection{External Reference Systems}

\paragraph{Language-Specific fastText Baselines.}
To determine whether joint family-level training improves over
conventional monolingual representation learning, we evaluate the
corresponding 300-dimensional embeddings from the official
157-language fastText release \citep{grave2018learning}. For each
language, we train a separate NER model using only its original-script
WikiANN training and validation data, without sharing labeled data across
languages. The downstream architecture and evaluation protocol match
those of the CTS and uroman conditions. Each monolingual experiment is
run with seeds 41, 42, and 43. Because this comparison changes both
embedding pretraining and downstream supervision, it serves as an
external end-to-end reference rather than a controlled ablation of script
unification.

\paragraph{CANINE-c Reference.}
A monolingual fastText comparison would cover only Turkish and Uyghur,
because these are the only languages in our selected UD data with both
training and validation splits. We therefore use CANINE-c, a multilingual
character-level Transformer encoder
\citep{clark-etal-2022-canine}, as a stronger external reference. We
fine-tune CANINE-c on the same pooled multilingual training and validation
splits, retaining the original orthographies, and evaluate all 19 treebanks
separately over five random seeds.

\section{Experimental Results}
% \subsection{Evaluation Protocol}
\label{sec:downstream section}
% We evaluated the learned representations on named entity recognition (NER) and part-of-speech (POS) tagging. For NER, we used WikiANN data for ten languages \citep{pan-etal-2017-cross,rahimi-etal-2019-massively}: \texttt{az}, \texttt{ba}, \texttt{cv}, \texttt{kk}, \texttt{ky}, \texttt{tk}, \texttt{tr}, \texttt{tt}, \texttt{ug}, and \texttt{uz}. For POS tagging, we used 19 UD v2.18 treebanks spanning seven languages \citep{nivre-etal-2020-universal}: \texttt{az}, \texttt{kk}, \texttt{ky}, \texttt{tr}, \texttt{tt}, \texttt{ug}, and \texttt{uz}. Sentence counts for the POS data splits are reported in Appendix~\ref{app:pos-data-splits}.

\begin{table}[t]
\centering
\scriptsize
\setlength{\tabcolsep}{2.5pt}
\renewcommand{\arraystretch}{0.9}
\resizebox{\columnwidth}{!}{%
\begin{tabular}{@{}lccrr@{}}
\toprule
Lang. & CTS & uroman & Raw $p$ & Holm $p$ \\
\midrule
\texttt{az} &
$\mathbf{0.9044} \pm 0.0042$ &
$0.9003 \pm 0.0051$ &
$0.2963$ & $1.0000$ \\
\texttt{ba} &
$0.8167 \pm 0.0146$ &
$\mathbf{0.8193} \pm 0.0267$ &
$0.8876$ & $1.0000$ \\
\texttt{cv} &
$\mathbf{0.8447} \pm 0.0117$ &
$0.8381 \pm 0.0205$ &
$0.4855$ & $1.0000$ \\
\texttt{kk} &
$\mathbf{0.8481} \pm 0.0055$ &
$0.8418 \pm 0.0052$ &
$0.0699$ & $0.6288$ \\
\texttt{ky} &
$0.6742 \pm 0.0393$ &
$\mathbf{0.7210} \pm 0.0259$ &
$0.0714$ & $0.6288$ \\
\texttt{tk} &
$0.7290 \pm 0.0150$ &
$\mathbf{0.7296} \pm 0.0353$ &
$0.9685$ & $1.0000$ \\
\texttt{tr} &
$\mathbf{0.9187} \pm 0.0012$ &
$0.9170 \pm 0.0009$ &
$0.0546$ & $0.5458$ \\
\texttt{tt} &
$\mathbf{0.9243} \pm 0.0045$ &
$0.9198 \pm 0.0058$ &
$0.1704$ & $1.0000$ \\
\texttt{ug} &
$\mathbf{0.6548} \pm 0.0189$ &
$0.6354 \pm 0.0210$ &
$0.2133$ & $1.0000$ \\
\texttt{uz} &
$0.9247 \pm 0.0024$ &
$\mathbf{0.9279} \pm 0.0060$ &
$0.1352$ & $0.9461$ \\
\midrule
Macro avg. &
$0.8240 \pm 0.0069$ &
$\mathbf{0.8250 \pm 0.0047}$ &
$0.8115$ & -- \\
\bottomrule
\end{tabular}%
}
\caption{WikiANN exact-match entity-level test $F_1$ (mean $\pm$ SD over five
seed-matched runs). Raw $p$ values use paired two-sided $t$-tests;
Holm correction is applied across languages
\citep{holm1979simple}. Bold indicates the higher mean.}
\label{tab:wikiann-results}
\end{table}

\label{sec:wikiann-results}
\subsection{NER: No Detectable Difference between CTS and uroman}

\begin{takeaway}
CTS and uroman perform nearly \textbf{identically on NER}.
\end{takeaway}

Table~\ref{tab:wikiann-results} reports exact-match entity-level test $F_1$
for the two representation conditions. CTS produces the higher numerical
mean for 6 of the 10 languages, whereas uroman produces the higher mean for
the remaining four. However, no per-language comparison is significant before
correction, and none remains
significant after Holm correction. The language-macro averages are also nearly identical, at 0.8240
for CTS and 0.8250 for uroman. The paired comparison of the seed-wise macro
averages shows no overall difference, $t(4)=-0.255$, $p=0.8115$.
Accordingly, the results provide \textbf{no evidence} that either representation is
\textbf{consistently superior} for WikiANN NER under the present experimental setup.

\subsection{Outperforming Language-Specific fastText Baselines}
\label{sec:wikiann-monolingual-fasttext}

\begin{takeaway}
Both CTS and uroman \textbf{substantially outperform} the \textbf{officially released} fastText baselines.
\end{takeaway}

\begin{table}[t]
\centering
\small
\resizebox{\columnwidth}{!}{%
\begin{tabular}{@{}lrrrr@{}}
\toprule
Lang. & Seed 41 & Seed 42 & Seed 43 & Mean {\scriptsize$\pm$ SD} \\
\midrule
\texttt{az} & 0.8770 & 0.8839 & 0.8717 & $0.8775$ {\scriptsize $\pm 0.0061$} \\
\texttt{ba} & 0.5644 & 0.4804 & 0.5789 & $0.5412$ {\scriptsize $\pm 0.0532$} \\
\texttt{cv} & 0.7043 & 0.7426 & 0.7100 & $0.7190$ {\scriptsize $\pm 0.0207$} \\
\texttt{kk} & 0.7421 & 0.7283 & 0.7388 & $0.7364$ {\scriptsize $\pm 0.0072$} \\
\texttt{ky} & 0.0529 & 0.3529 & 0.0473 & $0.1510$ {\scriptsize $\pm 0.1748$} \\
\texttt{tk} & 0.6304 & 0.6631 & 0.6162 & $0.6366$ {\scriptsize $\pm 0.0241$} \\
\texttt{tr} & 0.9064 & 0.9093 & 0.9069 & $0.9075$ {\scriptsize $\pm 0.0016$} \\
\texttt{tt} & 0.8447 & 0.8447 & 0.8406 & $0.8433$ {\scriptsize $\pm 0.0024$} \\
\texttt{ug} & 0.2013 & 0.6030 & 0.6400 & $0.4814$ {\scriptsize $\pm 0.2433$} \\
\texttt{uz} & 0.8647 & 0.8715 & 0.8721 & $0.8694$ {\scriptsize $\pm 0.0041$} \\
\midrule
Macro avg. & 0.6388 & 0.7080 & 0.6823 & $0.6763$ {\scriptsize $\pm 0.0350$} \\
\bottomrule
\end{tabular}%
}
\caption{WikiANN exact-match entity-level test $F_1$ obtained with the
language-specific models from the 157-language fastText release.}
\label{tab:wikiann-monolingual-fasttext}
\end{table}

The CTS- and uroman-based joint-training conditions both \textbf{substantially outperform} the monolingual fastText baselines at the language-macro and per-language levels. The gains are particularly pronounced for Bashkir and Kyrgyz, for which even the weakest run in each joint condition exceeds the strongest monolingual run by \textbf{more than 20\%} absolute $F_1$. More generally, for nine languages, the lowest-scoring CTS and uroman runs both surpass the best of the three monolingual runs. Uyghur is the sole exception, although its best monolingual score is still exceeded by four of the five CTS runs and two of the five uroman runs. The advantage also extends to higher-resourced Azerbaijani and Turkish: for both languages, the weakest run in each joint condition outperforms the strongest monolingual run. Thus, this targeted multilingual setup does not exhibit the high-resource degradation often associated with the ``curse of multilinguality'' in massively multilingual models \citep{conneau-etal-2020-unsupervised}. Although interference may occur in other settings, these results show that it is not inevitable when joint training is restricted to a closely related language family.

Because this comparison changes both embedding pretraining and downstream
supervision, it evaluates end-to-end training strategies rather than
isolating either source of improvement.

\subsection{POS: Effects Vary Across Languages}
\label{sec:ud-pos-results}

\begin{takeaway}
CTS and uroman each significantly outperform the other on several treebanks. Tables~\ref{tab:ud-pos-accuracy} and~\ref{tab:ud-pos-macro-f1} report
treebank-level test accuracy and macro-$F_1$, respectively. 
\end{takeaway}

\begin{table}[t]
\centering
\small
\setlength{\tabcolsep}{2pt}
\renewcommand{\arraystretch}{0.9}
\resizebox{\columnwidth}{!}{%
\begin{tabular}{@{}lccrr@{}}
\toprule
Treebank & CTS & uroman & Raw $p$ & Holm $p$ \\
\midrule
\texttt{az\_tuecl} &
$74.38 \pm 0.36$ & $\mathbf{77.19 \pm 0.80}$ &
$\mathbf{0.0011}$ & $\mathbf{0.0215}$ \\
\texttt{kk\_ktb} &
$\mathbf{75.54 \pm 0.50}$ & $73.81 \pm 0.73$ &
$\mathbf{0.0109}$ & $0.1632$ \\
\texttt{ky\_ktmu} &
$\mathbf{90.32 \pm 0.28}$ & $90.24 \pm 0.33$ &
$0.4781$ & $1.0000$ \\
\texttt{ky\_tuecl} &
$\mathbf{66.70 \pm 0.21}$ & $64.27 \pm 0.84$ &
$\mathbf{0.0034}$ & $0.0581$ \\
\texttt{tr\_atis} &
$\mathbf{97.88 \pm 0.14}$ & $97.84 \pm 0.22$ &
$0.7039$ & $1.0000$ \\
\texttt{tr\_boun} &
$90.36 \pm 0.32$ & $\mathbf{90.52 \pm 0.23}$ &
$\mathbf{0.0352}$ & $0.4483$ \\
\texttt{tr\_framenet} &
$\mathbf{95.42 \pm 0.31}$ & $95.31 \pm 0.22$ &
$0.5849$ & $1.0000$ \\
\texttt{tr\_gb} &
$\mathbf{89.54 \pm 0.24}$ & $89.47 \pm 0.26$ &
$0.7461$ & $1.0000$ \\
\texttt{tr\_imst} &
$89.23 \pm 0.09$ & $\mathbf{89.24 \pm 0.14}$ &
$0.8338$ & $1.0000$ \\
\texttt{tr\_kenet} &
$\mathbf{90.83 \pm 0.16}$ & $90.57 \pm 0.17$ &
$\mathbf{0.0345}$ & $0.4483$ \\
\texttt{tr\_penn} &
$\mathbf{93.11 \pm 0.10}$ & $92.98 \pm 0.22$ &
$0.2554$ & $1.0000$ \\
\texttt{tr\_pud} &
$83.64 \pm 0.18$ & $\mathbf{83.71 \pm 0.08}$ &
$0.5179$ & $1.0000$ \\
\texttt{tr\_tourism} &
$97.67 \pm 0.07$ & $\mathbf{97.69 \pm 0.10}$ &
$0.7408$ & $1.0000$ \\
\texttt{tr\_tuecl} &
$\mathbf{86.15 \pm 0.20}$ & $85.47 \pm 0.17$ &
$\mathbf{0.0020}$ & $\mathbf{0.0354}$ \\
\texttt{tt\_nmctt} &
$69.72 \pm 0.37$ & $\mathbf{71.55 \pm 0.58}$ &
$\mathbf{0.0046}$ & $0.0733$ \\
\texttt{ug\_udt} &
$\mathbf{87.28 \pm 0.58}$ & $86.13 \pm 0.23$ &
$\mathbf{0.0167}$ & $0.2343$ \\
\texttt{uz\_tuecl} &
$\mathbf{76.85 \pm 0.95}$ & $75.40 \pm 1.32$ &
$0.1942$ & $1.0000$ \\
\texttt{uz\_ut} &
$\mathbf{82.54 \pm 1.01}$ & $81.57 \pm 0.68$ &
$0.1800$ & $1.0000$ \\
\texttt{uz\_uzudt} &
$\mathbf{86.85 \pm 0.70}$ & $85.92 \pm 0.39$ &
$0.0769$ & $0.8456$ \\
\midrule
Macro avg. &
$\mathbf{85.47 \pm 0.20}$ & $85.20 \pm 0.18$ &
$0.0783$ & -- \\
\bottomrule
\end{tabular}%
}
\caption{UD POS test accuracy (mean $\pm$ sample standard deviation over
five seed-matched runs). The macro average weights all 19 treebanks equally
and is computed separately for each seed. Boldface marks the higher mean;
bold $p$ values indicate $p<0.05$.}
\label{tab:ud-pos-accuracy}
\end{table}

\begin{table}[t]
\centering
\small
\setlength{\tabcolsep}{2pt}
\renewcommand{\arraystretch}{0.9}
\resizebox{\columnwidth}{!}{%
\begin{tabular}{@{}lccrr@{}}
\toprule
Treebank & CTS & uroman & Raw $p$ & Holm $p$ \\
\midrule
\texttt{az\_tuecl} &
$53.45 \pm 1.04$ & $\mathbf{56.89 \pm 1.70}$ &
$\mathbf{0.0124}$ & $0.1977$ \\
\texttt{kk\_ktb} &
$\mathbf{43.04 \pm 1.20}$ & $40.76 \pm 2.13$ &
$0.1534$ & $1.0000$ \\
\texttt{ky\_ktmu} &
$66.66 \pm 1.08$ & $\mathbf{72.60 \pm 1.68}$ &
$\mathbf{0.0012}$ & $\mathbf{0.0215}$ \\
\texttt{ky\_tuecl} &
$\mathbf{35.43 \pm 1.18}$ & $34.24 \pm 1.62$ &
$0.0947$ & $1.0000$ \\
\texttt{tr\_atis} &
$96.71 \pm 3.42$ & $\mathbf{96.80 \pm 3.29}$ &
$0.5607$ & $1.0000$ \\
\texttt{tr\_boun} &
$80.84 \pm 0.92$ & $\mathbf{81.07 \pm 0.73}$ &
$0.5765$ & $1.0000$ \\
\texttt{tr\_framenet} &
$\mathbf{94.93 \pm 0.45}$ & $93.75 \pm 1.34$ &
$0.0811$ & $1.0000$ \\
\texttt{tr\_gb} &
$69.98 \pm 0.58$ & $\mathbf{70.09 \pm 0.13}$ &
$0.7302$ & $1.0000$ \\
\texttt{tr\_imst} &
$79.53 \pm 0.88$ & $\mathbf{80.45 \pm 1.84}$ &
$0.2076$ & $1.0000$ \\
\texttt{tr\_kenet} &
$86.73 \pm 0.60$ & $\mathbf{86.91 \pm 0.47}$ &
$0.3412$ & $1.0000$ \\
\texttt{tr\_penn} &
$\mathbf{91.71 \pm 1.80}$ & $91.49 \pm 1.79$ &
$0.8156$ & $1.0000$ \\
\texttt{tr\_pud} &
$\mathbf{67.19 \pm 0.22}$ & $66.95 \pm 0.23$ &
$0.0801$ & $1.0000$ \\
\texttt{tr\_tourism} &
$\mathbf{91.91 \pm 0.76}$ & $91.78 \pm 0.56$ &
$0.5023$ & $1.0000$ \\
\texttt{tr\_tuecl} &
$\mathbf{77.31 \pm 0.58}$ & $76.59 \pm 1.09$ &
$0.1257$ & $1.0000$ \\
\texttt{tt\_nmctt} &
$36.65 \pm 0.93$ & $\mathbf{43.29 \pm 1.26}$ &
$\mathbf{0.0004}$ & $\mathbf{0.0077}$ \\
\texttt{ug\_udt} &
$\mathbf{76.94 \pm 1.23}$ & $73.22 \pm 1.02$ &
$\mathbf{0.0061}$ & $0.1039$ \\
\texttt{uz\_tuecl} &
$\mathbf{60.44 \pm 4.91}$ & $54.99 \pm 2.77$ &
$0.1176$ & $1.0000$ \\
\texttt{uz\_ut} &
$\mathbf{56.52 \pm 2.56}$ & $55.18 \pm 1.76$ &
$0.5220$ & $1.0000$ \\
\texttt{uz\_uzudt} &
$64.05 \pm 3.60$ & $\mathbf{68.26 \pm 4.06}$ &
$0.0554$ & $0.8305$ \\
\midrule
Macro avg. &
$70.00 \pm 0.58$ & $\mathbf{70.28 \pm 0.60}$ &
$0.4888$ & -- \\
\bottomrule
\end{tabular}%
}
\caption{UD POS test macro-$F_1$ (mean $\pm$ sample standard deviation over
five seed-matched runs). The macro average weights all 19 treebanks equally
and is computed separately for each seed. Boldface marks the higher mean;
bold $p$ values indicate $p<0.05$.}
\label{tab:ud-pos-macro-f1}
\end{table}

\paragraph{Accuracy.}~ CTS attains higher mean accuracy on 13 of the 19 treebanks, spanning 5 languages, whereas uroman does so on the remaining 6 treebanks, spanning 3 languages. At the uncorrected $p<0.05$ level, CTS yields significantly higher accuracy on 5 treebanks, whereas uroman does so on 3. After Holm correction, only the uroman advantage on \texttt{az\_tuecl} and the CTS advantage on \texttt{tr\_tuecl} remain significant. Complete per-treebank results are reported in Table~\ref{tab:ud-pos-accuracy}.

\paragraph{Macro-$F_1$.}~ CTS attains higher mean macro-$F_1$ on 10 treebanks spanning 5 languages, whereas uroman does so on the remaining 9 treebanks, also spanning 5 languages. At the uncorrected $p<0.05$ level, 4 comparisons are significant: uroman yields significantly higher macro-$F_1$ on 3 treebanks, whereas CTS does so only on \texttt{ug\_udt}. After Holm correction, the uroman advantages on \texttt{ky\_ktmu} and \texttt{tt\_nmctt} remain significant, whereas no CTS advantage does. Complete per-treebank results are reported in Table~\ref{tab:ud-pos-macro-f1}.

\subsection{CANINE-c Comparison}
\label{sec:ud-pos-canine}

CTS numerically exceeds CANINE-c on 7/8 treebanks in
accuracy/macro-$F_1$, versus 5/7 for uroman; 6/2 and 4/1 of these
advantages, respectively, are significant before correction. After Holm
correction over 19 treebanks per representation--metric family, both
retain significant accuracy advantages on three treebanks, while CTS
also retains one macro-$F_1$ advantage. Full results appear in
Appendix~\ref{app:canine-significance}.

% \subsection{Comparison with CANINE}
% \label{sec:ud-pos-canine}

% % \paragraph{Treebank-specific Advantages.}

% CTS numerically exceeds CANINE-c on 7 treebanks in accuracy and 8 in
% macro-$F_1$, while uroman does so on 5 and 7 treebanks, respectively.
% At the uncorrected $p<0.05$ level, CTS significantly outperforms
% CANINE-c on 6 treebanks in accuracy and 2 in macro-$F_1$, while uroman
% does so on 4 and 1 treebanks, respectively. After Holm correction across
% the 19 treebanks within each representation--metric family, CTS retains
% accuracy advantages on \texttt{tr\_gb}, \texttt{tr\_pud}, and
% \texttt{uz\_uzudt}, as well as a macro-$F_1$ advantage on
% \texttt{ug\_udt}. Uroman retains accuracy advantages on
% \texttt{tr\_gb}, \texttt{tr\_pud}, and \texttt{uz\_uzudt}, whereas
% none of its macro-$F_1$ advantages survives correction. Full statistical
% results are reported in Appendix~\ref{app:canine-significance}.
% Several of these advantages are significant at the uncorrected
% $p<0.05$ level, and a smaller subset remains significant after Holm
% correction. Full treebank-level results are reported in
% Table~\ref{tab:ud-pos-canine}.

CANINE-c achieves the highest treebank-macro averages
(86.20/72.39 accuracy/macro-$F_1$), versus 85.47/70.00 for CTS and
85.20/70.28 for uroman. This suggests that the larger character-level
pretrained encoder performs better overall, although the substantially
simpler fastText systems match or exceed it on some treebanks. Because the
systems differ in architecture, pretraining, and input representation,
CANINE-c serves as an external reference rather than a controlled ablation.
% \paragraph{Overall Comparison.}
% CANINE-c achieves the highest treebank-macro averages, with 86.20 accuracy
% and 72.39 macro-$F_1$, compared with 85.47 and 70.00 for CTS and 85.20 and
% 70.28 for uroman. Thus, the larger character-level pretrained encoder performs better
% overall, while the substantially simpler fastText-based systems can still
% match or outperform it in particular evaluation settings.
% % \paragraph{Scope.}
% Because CANINE-c differs from our systems in architecture, pretraining, and
% input representation, this comparison should be interpreted as an external
% reference rather than a controlled ablation of script unification.

\begin{table}[t]
\centering
\small
\setlength{\tabcolsep}{4pt}
\renewcommand{\arraystretch}{0.9}
\begin{tabular}{@{}lcc@{}}
\toprule
Treebank & Accuracy & Macro-$F_1$ \\
\midrule
\texttt{az\_tuecl}       & $78.49 \pm 0.62$ & $61.59 \pm 1.24$ \\
\texttt{kk\_ktb}         & $77.78 \pm 0.39$ & $50.01 \pm 0.68$ \\
\texttt{ky\_ktmu}        & $90.99 \pm 0.11$ & $76.22 \pm 0.98$ \\
\texttt{ky\_tuecl}       & $69.38 \pm 0.87$ & $39.78 \pm 1.39$ \\
\texttt{tr\_atis}        & $98.25 \pm 0.13$ & $98.32 \pm 0.18$ \\
\texttt{tr\_boun}        & $89.80 \pm 0.28$ & $80.73 \pm 0.72$ \\
\texttt{tr\_framenet}    & $95.63 \pm 0.30$ & $95.04 \pm 0.78$ \\
\texttt{tr\_gb}          & $88.53 \pm 0.06$ & $69.52 \pm 0.38$ \\
\texttt{tr\_imst}        & $89.85 \pm 0.28$ & $81.40 \pm 0.95$ \\
\texttt{tr\_kenet}       & $91.19 \pm 0.22$ & $86.98 \pm 0.28$ \\
\texttt{tr\_penn}        & $93.03 \pm 0.16$ & $90.40 \pm 0.39$ \\
\texttt{tr\_pud}         & $82.90 \pm 0.20$ & $66.74 \pm 0.25$ \\
\texttt{tr\_tourism}     & $97.86 \pm 0.10$ & $93.79 \pm 0.66$ \\
\texttt{tr\_tuecl}       & $86.35 \pm 0.78$ & $75.89 \pm 0.77$ \\
\texttt{tt\_nmctt}       & $78.76 \pm 0.81$ & $55.36 \pm 1.95$ \\
\texttt{ug\_udt}         & $86.23 \pm 0.27$ & $70.67 \pm 0.81$ \\
\texttt{uz\_tuecl}       & $74.82 \pm 0.50$ & $59.12 \pm 1.23$ \\
\texttt{uz\_ut}          & $83.60 \pm 0.37$ & $60.24 \pm 0.78$ \\
\texttt{uz\_uzudt}       & $84.43 \pm 0.34$ & $63.55 \pm 1.06$ \\
\midrule
Macro avg.                & $86.20 \pm 0.11$ & $72.39 \pm 0.24$ \\
\bottomrule
\end{tabular}
\caption{Original-script UD POS test performance of CANINE-c
(mean $\pm$ sample standard deviation over five runs). The macro average
weights all 19 treebanks equally and is computed separately for each seed.}
\label{tab:ud-pos-canine}
\end{table}

\section{Discussion}
\label{sec:ud-pos-discussion}

\subsection{Benefits of Family-Level Joint Training}
One plausible explanation for CTS and uroman outperforming monolingual fastText models on higher-resourced Turkish and Azerbaijani is that CTS and uroman make writing systems more comparable across the family, thereby increasing the cross-lingual character $n$-gram overlap available to fastText. Because this representational sharing aligns with genuine linguistic similarities, including cognate vocabulary and broadly shared agglutinative morphology, data from other Turkic languages may provide structured variation rather than unrelated noise. Under this interpretation, the pooled corpus forms a family-level continuum of mutually informative varieties, and joint training functions as structured data augmentation that improves robustness and generalization even for higher-resource languages.

\subsection{POS Mechanistic Hypothesis}
We examine \texttt{az\_tuecl}, \texttt{tt\_nmctt}, and \texttt{ug\_udt}, the only three treebanks for which one transliteration outperforms the other on both accuracy and macro-$F_1$ at the uncorrected $p<0.05$ level. Neither \texttt{az\_tuecl} nor \texttt{tt\_nmctt} provides training or validation data, so their test performance depends entirely on zero-shot transfer from the other languages.

Because fastText constructs word representations from character $n$-grams, we hypothesize that uroman's stronger zero-shot performance on Azerbaijani and Tatar results from greater character $n$-gram overlap between their test tokens and the pooled training data from the other languages. Uyghur, by contrast, provides both training and validation data. Its performance may therefore depend on two forms of subword sharing: cross-lingual overlap between the Uyghur test set and the non-Uyghur training data, and within-language overlap between the Uyghur test and training sets.

To evaluate these hypotheses, we measure the literal character 2--5-gram coverage of each target test set by the relevant POS training tokens under CTS and uroman transliteration. Token-mean coverage measures the average proportion of covered $n$-grams within a target token; occurrence coverage weights each target $n$-gram occurrence equally; and type coverage weights each distinct target $n$-gram equally. Full definitions are provided in Appendix~\ref{app:ngram-coverage-metrics}. These measurements allow us to test whether the observed performance differences align with the cross-lingual and within-language subword sharing induced by each transliteration.

\subsection{Analysis of Azerbaijani and Tatar}
We measure cross-lingual coverage by using the pooled POS training splits of all other languages as the reference corpus and calculating how much of the character $n$-gram structure in the target language's test tokens is covered by that corpus. Relative to CTS, uroman increases Azerbaijani accuracy by 2.81 percentage points and macro-$F_1$ by 3.44 points; for Tatar, the corresponding improvements are 1.83 and 6.63 points. Table~\ref{tab:ud-pos-az-tt-ngram-coverage} shows the same pattern in the proposed mechanism: for both target languages, uroman produces higher coverage by the other languages' pooled training splits at every $n$-gram order and under all three coverage measures. Thus, in these two zero-shot settings, uroman creates greater task-relevant subword sharing between each target-language test set and the available training data, which coincides with better POS transfer.
\begin{table}[t]
\centering
\footnotesize
\setlength{\tabcolsep}{1.8pt}
\renewcommand{\arraystretch}{0.9}
\begin{tabular}{@{}lllrrrrr@{}}
\toprule
Target & Rep. & Measure & 2 & 3 & 4 & 5 & 2--5 \\
\midrule
\texttt{az} & CTS &
Tok.      & 92.21 & 82.35 & 63.83 & 50.09 & 79.52 \\
\texttt{az} & CTS &
Occ.      & 91.26 & 78.16 & 62.68 & 46.96 & 73.76 \\
\texttt{az} & CTS &
Type      & 91.55 & 75.67 & 58.49 & 41.07 & 60.89 \\
\addlinespace[1pt]
\texttt{az} & uroman &
Tok.      & 99.96 & 97.79 & 89.00 & 76.15 & 94.29 \\
\texttt{az} & uroman &
Occ.      & 99.93 & 97.07 & 88.61 & 75.21 & 92.31 \\
\texttt{az} & uroman &
Type      & 99.47 & 95.36 & 86.37 & 71.35 & 84.49 \\
\midrule
\texttt{tt} & CTS &
Tok.      & 93.22 & 81.66 & 58.86 & 37.03 & 76.12 \\
\texttt{tt} & CTS &
Occ.      & 92.41 & 78.91 & 59.13 & 38.54 & 71.36 \\
\texttt{tt} & CTS &
Type      & 91.24 & 76.30 & 54.86 & 34.42 & 54.42 \\
\addlinespace[1pt]
\texttt{tt} & uroman &
Tok.      & 99.95 & 98.67 & 84.41 & 53.68 & 89.93 \\
\texttt{tt} & uroman &
Occ.      & 99.94 & 98.52 & 84.34 & 54.47 & 87.49 \\
\texttt{tt} & uroman &
Type      & 98.78 & 95.49 & 76.49 & 47.84 & 69.76 \\
\bottomrule
\end{tabular}
\caption{Literal cross-lingual character $n$-gram coverage (\%) of the
transliterated Azerbaijani (\texttt{az}) and Tatar (\texttt{tt}) test tokens
by the pooled training splits of the other treebanks.}
\label{tab:ud-pos-az-tt-ngram-coverage}
\end{table}

\subsection{Analysis of Uyghur}
Unlike Azerbaijani and Tatar, \texttt{ug\_udt} provides its own training data.
We therefore asked whether CTS's advantage on Uyghur reflects greater literal
$n$-gram overlap with (i) the non-UG training data, which could facilitate
cross-lingual transfer, or (ii) the UG-only training data, which could
facilitate within-language training. Table~\ref{tab:ud-pos-ug-ngram-coverage}
reports both comparisons.

\begin{table}[t]
\centering
\footnotesize
\setlength{\tabcolsep}{2pt}
\renewcommand{\arraystretch}{0.9}
\resizebox{\columnwidth}{!}{%
\begin{tabular}{@{}lllrrrrr@{}}
\toprule
Pool & Rep. & Measure. & 2 & 3 & 4 & 5 & 2--5 \\
\midrule
Non-UG & CTS &
Tok.  & 97.15 & 92.84 & 74.95 & 49.17 & 85.62 \\
&     &
Occ.  & 96.66 & 91.16 & 73.04 & 44.44 & 80.62 \\
&     &
Type  & 86.33 & 78.26 & 58.90 & 33.22 & 50.62 \\
\addlinespace[1pt]
& uroman &
Tok.  & 99.53 & 91.33 & 58.67 & 30.87 & 79.62 \\
&        &
Occ.  & 99.41 & 89.76 & 57.75 & 29.36 & 73.84 \\
&        &
Type  & 88.11 & 72.62 & 49.45 & 24.49 & 39.33 \\
\midrule
UG only & CTS &
Tok.  & 99.77 & 98.63 & 92.11 & 82.68 & 95.70 \\
&     &
Occ.  & 99.65 & 98.08 & 90.94 & 79.79 & 93.75 \\
&     &
Type  & 88.23 & 83.70 & 71.92 & 59.44 & 68.35 \\
\addlinespace[1pt]
& uroman &
Tok.  & 99.83 & 99.35 & 96.78 & 91.34 & 97.86 \\
&        &
Occ.  & 99.74 & 99.10 & 96.44 & 90.42 & 97.02 \\
&        &
Type  & 86.16 & 83.61 & 79.37 & 71.08 & 75.46 \\
\bottomrule
\end{tabular}%
}
\caption{Literal character $n$-gram coverage (\%) of the transliterated
\texttt{ug\_udt} test tokens by the non-UG and UG-only training pools.}
\label{tab:ud-pos-ug-ngram-coverage}
\end{table}

\paragraph{n-gram Coverage Evidence.} Against the non-UG training pool, CTS leads uroman in pooled 2--5-gram
token-mean, occurrence, and type coverage by 6.00, 6.78, and 11.29 percentage
points, respectively. In contrast, against the UG-only pool, uroman leads by
2.16, 3.27, and 7.11 points. The pronounced non-UG coverage advantage of CTS
may therefore help explain its significantly higher Uyghur accuracy and macro-$F_1$ in
Tables~\ref{tab:ud-pos-accuracy} and~\ref{tab:ud-pos-macro-f1}; the coverage results do not
support greater within-Uyghur overlap as the explanation.

\paragraph{Further POS Ablation Study.} To test whether this pattern carries over to downstream performance, we
reran the POS model under two controlled regimes. The non-UG regime excluded
Uyghur from both training and validation, whereas the UG-only regime used
only Uyghur training and validation data. Both experiments used the same CTS and uroman settings, model architecture, training procedure, hyperparameters, and five seeds as
Section~\ref{sec:script-unification}.

\begin{table}[t]
\centering
\footnotesize
\setlength{\tabcolsep}{2.5pt}
\renewcommand{\arraystretch}{0.9}
\resizebox{\columnwidth}{!}{%
\begin{tabular}{@{}llccrr@{}}
\toprule
Regime & Metric & CTS & uroman & Raw $p$ & Holm $p$ \\
\midrule
Non-UG & Accuracy &
$\mathbf{76.65 \pm 0.37}$ & $65.30 \pm 1.29$ &
$\mathbf{0.0001}$ & $\mathbf{0.0002}$ \\
& Macro-$F_1$ &
$\mathbf{43.82 \pm 1.92}$ & $24.01 \pm 0.83$ &
$\mathbf{0.0001}$ & $\mathbf{0.0002}$ \\
\addlinespace[1pt]
UG only & Accuracy &
$\mathbf{89.14 \pm 0.35}$ & $88.37 \pm 0.13$ &
$\mathbf{0.0138}$ & $\mathbf{0.0275}$ \\
& Macro-$F_1$ &
$\mathbf{77.84 \pm 0.83}$ & $76.95 \pm 0.53$ &
$0.1375$ & $0.1375$ \\
\bottomrule
\end{tabular}%
}
\caption{Uyghur POS test scores (%; mean $\pm$ sample standard deviation
over five seed-matched runs). Raw $p$ values are from two-sided paired
$t$-tests; Holm correction is applied jointly across the four comparisons.
Boldface marks the higher mean and $p<0.05$.}
\label{tab:ud-pos-ug-data-ablation}
\end{table}

Table~\ref{tab:ud-pos-ug-data-ablation} provides strong evidence that CTS's principal advantage on Uyghur comes from cross-lingual transfer. Under non-UG POS supervision, CTS significantly outperforms uroman in both accuracy and macro-$F_1$. The gaps are substantially smaller when Uyghur supervision is included in the multilingual setting of Tables~\ref{tab:ud-pos-accuracy} and~\ref{tab:ud-pos-macro-f1}, and narrow further under UG-only POS supervision. The reduction from the non-UG to the UG-only regime is noticeable for both metrics. Together with Table~\ref{tab:ud-pos-ug-ngram-coverage}, this pattern suggests that CTS transfers more effectively from the other Turkic languages, whereas uroman's higher within-Uyghur train--test coverage helps it considerably when Uyghur supervision is available.

CTS nevertheless retains a small mean advantage under UG-only POS supervision, although only the accuracy difference is significant. One tentative explanation is that CTS handles Uyghur, the only Perso-Arabic-script language in these experiments, more effectively. Its transliteration may yield greater overlap with other Turkic languages and/or fewer collisions between distinct Uyghur forms, whereas uroman's high within-language coverage may partly reflect reduced distinctions among romanized forms. Such effects could influence both fastText pretraining and downstream POS learning. Because the residual advantage is small and the UG-only ablation retains the jointly pretrained frozen fastText models, its source cannot be isolated here; this explanation therefore remains preliminary.

\paragraph{Overall takeaway.}
The preferred representation depends on the language and supervision regime: cross-lingual coverage is central to zero-shot transfer, whereas within-language coverage can narrow differences when target supervision is available.

\section{Related Work}
\citet{khatri2020study} trained a single fastText model on eleven Indic Wikipedia corpora transliterated into Devanagari and reported stronger bilingual lexicon induction than post-hoc embedding alignment for most language pairs. \citet{moosa-etal-2023-transliteration} found that common-script language modeling particularly benefited lower-resource Indic languages. In multilingual machine translation, \citet{nguyen-chiang-2017-transfer} and \citet{sun-etal-2022-alternative} showed that shared or alternative scripts can increase useful subword sharing among related languages, including Turkic languages. Surveys  characterize transliteration as effective but task- and language-dependent \citep{jayakumar-etal-2026-scripts}. Unlike prior Turkic studies centered on supervised translation or restricted transfer settings, we  compare universal and family-specific normalization in a fully mixed, language-ID-free eleven-language representation model.

\section{Conclusion}

We compared universal and family-specific script unification for joint modeling of 11 Turkic languages. CTS and uroman perform similarly on WikiANN NER, while both substantially outperform the official monolingual fastText baselines. On UD POS tagging, neither representation is consistently superior: uroman performs better in the Azerbaijani and Tatar zero-shot settings, whereas CTS performs better in Uyghur. These differences are associated primarily with cross-lingual character $n$-gram coverage, while target-language supervision increases the importance of within-language overlap and narrows representation differences. CANINE-c achieves stronger overall POS averages, but the much simpler fastText-based systems remain competitive on several treebanks. Our results therefore suggest that a unified writing system should be selected according to the language, the supervision regime, and the subword overlap it induces, rather than treated as a universally optimal representation.

\section*{Limitations}

Our conclusions are limited to fastText and the WikiANN NER and UD POS
settings. Although CTS and uroman show differences in several uncorrected
POS comparisons for Azerbaijani, Tatar, and Uyghur, most NER and POS
comparisons do not reach significance. This may partly reflect the limited
statistical power provided by five runs per condition. It may also reflect
our use of large pooled Wikipedia corpora: extensive multilingual
pretraining, together with fastText's character $n$-gram mechanism, may
reduce sensitivity to subtle differences between script-unification
methods.

Our analysis also does not address learned subword tokenization in modern
Transformer models. Many multilingual systems rely on BPE-like tokenizers
whose vocabularies and merge operations are estimated from surface string
statistics. Script unification may reduce script-specific vocabulary
fragmentation, increase the reuse of subword units across related
languages, and lower tokenization fertility or sequence length. At the
same time, small orthographic differences between CTS and uroman can alter
merge statistics and cause related forms to be segmented into different
subword sequences. It therefore remains unclear which representation
provides more effective vocabulary allocation and cross-lingual subword
sharing under a fixed tokenizer vocabulary and training corpus.

Moreover, our two-stage pipeline---unsupervised embedding pretraining
followed by supervised downstream training---does not reveal whether
observed differences arise during representation learning, downstream
learning, or their interaction. Future work could use smaller training
corpora, more random seeds, and more demanding transfer settings. It could
also train matched BPE, WordPiece, or Unigram tokenizers on the CTS and
uroman corpora and compare vocabulary overlap, tokenization fertility,
sequence length, and downstream Transformer performance. A single-stage
alternative would train character-level NER or POS models from scratch
directly on WikiANN or UD supervision. A particularly stringent design
would use labeled data from only one source language, such as Turkish, and
evaluate zero-shot transfer to the remaining Turkic languages, thereby
isolating the relative effects of CTS and uroman more clearly.

% Bibliography entries are in references.bib.
\bibliography{custom}

\appendix

\section{Detailed Data Processing}
\label{app:data-processing}

\subsection{Shared Data Processing}

\begin{figure*}[t]
  \centering
  \begin{tikzpicture}[
      box/.style={
        draw,
        rounded corners=1.5pt,
        align=center,
        font=\scriptsize,
        inner xsep=2.5pt,
        inner ysep=2.5pt
      },
      arrow/.style={-{Latex[length=1.5mm]}, thin}
    ]
    \node[box, text width=14mm] (raw) {Raw data};
    \node[box, text width=39mm, right=4mm of raw] (pre) {
      \textbf{Shared preprocessing}\\
      NFC $\rightarrow$ remove trailing CR/LF\\
      $\rightarrow$ normalize apostrophes to U+0027
    };
    \node[box, text width=25mm, right=5mm of pre, yshift=6mm] (cts) {
      TurkicNLP CTS
    };
    \node[box, text width=25mm, right=5mm of pre, yshift=-6mm] (uroman) {
      uroman
    };
    \node[box, text width=38mm, right=5mm of cts, yshift=-6mm] (post) {
      \textbf{Shared postprocessing}\\
      NFC $\rightarrow$ UAX~\#29 tokenization\\
      $\rightarrow$ single-space serialization
    };
    \node[box, text width=27mm, right=4mm of post] (files) {
      11 CTS and 11 uroman text files
    };

    \draw[arrow] (raw) -- (pre);
    \draw[arrow] (pre.east) -- (cts.west);
    \draw[arrow] (pre.east) -- (uroman.west);
    \draw[arrow] (cts.east) -- (post.north west);
    \draw[arrow] (uroman.east) -- (post.south west);
    \draw[arrow] (post) -- (files);
  \end{tikzpicture}
  \caption{Data-level script unification and normalization. The CTS and
  uroman conditions differ only in the representation-specific branch.}
  \label{fig:data-level-normalization}
\end{figure*}
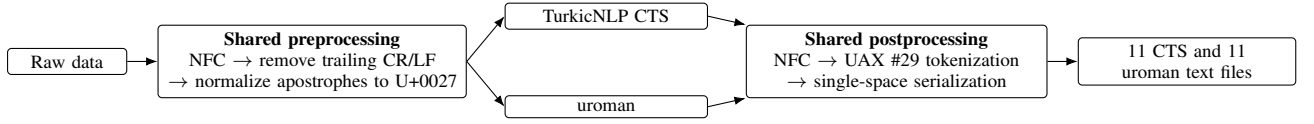

For each of the 11 Turkic-language Wikipedia editions, we constructed
parallel CTS and uroman corpora from the same January~1, 2026 Wikimedia dump
snapshot. As Figure~\ref{fig:data-level-normalization} shows, the two
conditions shared several processing steps; only the script
transliteration branch differed. Before branching, each raw data was
normalized to NFC, stripped of trailing carriage-return and newline
characters, and canonicalized by mapping ten apostrophe-like characters to the ASCII apostrophe (U+0027). The ten source characters and their Unicode code points are listed in
Appendix~\ref{app:apostrophe-normalization}. 

After the representation-specific transliterations (which is the use of uroman and TurkicNLP CTS transliteration), the CTS and uroman branches rejoined a shared final processing pipeline. The data was again normalized to NFC and tokenized according to the Unicode word-boundary rules in UAX~\#29. Tokens that remained nonempty after stripping surrounding whitespace were joined with single spaces and written to plain-text files. After all data processing steps above, we obtained matched CTS and uroman Wikipedia corpora for all eleven Turkic languages.

\subsection{CTS-Specific Processing}
\label{app:cts-specific-processing}

We used the \texttt{Transliterator} interface in TurkicNLP \citep{hakimov2026turkicnlpnlptoolkit}. A transliterator is instantiated with a language code, a declared source script, and a target script. For all CTS transliterations, the target was \texttt{Script.COMMON\_TURKIC}. Because a Wikipedia edition may contain more than one orthography of its primary language and terms borrowed from other languages when describing related concepts, to maximize transliterate the raw script into the shared CTS representation space, we applied four transliterators sequentially:
\[
\begin{aligned}
D_1 &= T_1(D_0), && T_1:\mathrm{Latin}\rightarrow\mathrm{CTS},\\
D_2 &= T_2(D_1), && T_2:\mathrm{Cyrillic}\rightarrow\mathrm{CTS},\\
D_3 &= T_3(D_2), && T_3:\mathrm{PersoArabic}\rightarrow\mathrm{CTS},\\
D_4 &= T_4(D_3), && T_4:\mathrm{PersoArabic}\rightarrow\mathrm{CTS},
\end{aligned}
\]

where \(D_0\) denotes the shared-processed input. Characters that are not recognized by a transliterator are passed through unchanged and remain available to subsequent stages.

Whenever TurkicNLP provided a source-language-specific transliteration for a given script, we used it. When no such transliterator was available for a given script of a source language, we used a deliberately chosen proxy transliteration rather than omitting that script stage to maximize the transliteration of script into a shared CTS representation space. Table \ref{tab:cts-transliterators} gives the complete configuration.

\begin{table}[t]
\centering
\footnotesize
\setlength{\tabcolsep}{3.5pt}
\renewcommand{\arraystretch}{0.9}
\begin{tabular}{@{}lcccc@{}}
\toprule
Wiki & $T_1$ & $T_2$ & $T_3$ & $T_4$ \\
\midrule
\texttt{az}  & \texttt{aze} & \texttt{aze} & \texttt{azb} & \texttt{uig} \\
\texttt{ba}  & \texttt{kaa}\textsuperscript{*}
             & \texttt{bak} & \texttt{uig} & \texttt{azb} \\
\texttt{cv}  & \texttt{kaa}\textsuperscript{*}
             & \texttt{chv} & \texttt{uig} & \texttt{azb} \\
\texttt{kaa} & \texttt{kaa} & \texttt{kaa} & \texttt{uig} & \texttt{azb} \\
\texttt{kk}  & \texttt{kaz} & \texttt{kaz} & \texttt{uig} & \texttt{azb} \\
\texttt{ky}  & \texttt{kaa}\textsuperscript{*}
             & \texttt{kir} & \texttt{uig} & \texttt{azb} \\
\texttt{tt}  & \texttt{tat} & \texttt{tat} & \texttt{uig} & \texttt{azb} \\
\texttt{tk}  & \texttt{tuk} & \texttt{tuk} & \texttt{uig} & \texttt{azb} \\
\texttt{tr}  & \texttt{tur} & \texttt{kaz}\textsuperscript{*}
             & \texttt{uig} & \texttt{azb} \\
\texttt{ug}  & \texttt{uig} & \texttt{uig} & \texttt{uig} & \texttt{azb} \\
\texttt{uz}  & \texttt{uzb} & \texttt{uzb} & \texttt{uig} & \texttt{azb} \\
\bottomrule
\end{tabular}

\caption{TurkicNLP configurations for the four CTS stages. All entries
target \texttt{Script.COMMON\_TURKIC}. An asterisk marks a proxy mapping used when no
source-language-specific transliteration is available.}
\label{tab:cts-transliterators}
\end{table}

For Bashkir, Chuvash, and Kyrgyz, TurkicNLP does not provide a Latin-to-CTS transliterator, so we used its Karakalpak (\texttt{kaa}) Latin-to-CTS transliterator as \(T_1\). The choice was motivated by the comparatively conservative nature of this transliteration: most basic Latin letters are preserved, while a limited set of common sequences and CTS-relevant characters are normalized (e.g., \texttt{sh} \(\rightarrow\) \texttt{ş} and \texttt{ch} \(\rightarrow\) \texttt{ç}).

Conversely, TurkicNLP provides no Turkish Cyrillic-to-CTS transliterator because Turkish has no standard Cyrillic orthography. We therefore used the Kazakh (\texttt{kaz}) Cyrillic-to-CTS transliterator as \(T_2\) for Turkish. This transliteration covers the Russian Cyrillic base together with several widely encountered Turkic Cyrillic extensions.

% \paragraph{Ordering of the Latin and Cyrillic stages.}
% The Latin stage \(T_1\) precedes the Cyrillic stage \(T_2\) intentionally. TurkicNLP's transliteration rules operate on matching strings rather than first verifying the script identity of every token. If Cyrillic-to-CTS transliteration were performed first, its Latin-script CTS output would subsequently become eligible for the Latin-to-CTS rules. This could transliterate newly produced CTS strings a second time and create cascaded substitution errors. In the chosen order, \(T_1\) converts pre-existing Latin material, after which \(T_2\) converts Cyrillic material while leaving the Latin CTS output from \(T_1\) unchanged.

\paragraph{Two-stage Perso-Arabic transliteration.}
We used two Perso-Arabic-to-CTS transliterators to increase character coverage. For every language except Azerbaijani, \(T_3\) was the Uyghur (\texttt{uig}) transliterator and \(T_4\) was the South Azerbaijani (\texttt{azb}) transliterator. Modern Uyghur Perso-Arabic orthography writes vowels obligatorily with dedicated letters, whereas the South Azerbaijani Perso-Arabic transliteration is abjad-based and does not consistently encode short vowels \citep{kontovas2021readinguyghur,loc2011azerbaijaniromanization}. Applying the Uyghur transliterator first therefore provides broad initial coverage, while the South Azerbaijani transliterator acts as a coverage-oriented fallback. For the Azerbaijani Wikipedia, we reversed this order: the closely related South Azerbaijani transliterator was applied first to maximize the accuracy of likely South Azerbaijani material, and the Uyghur transliterator was then used as a coverage-oriented fallback.

This procedure is not a token-level, language-specific transliteration system. Once \(T_3\) transliterates a Perso-Arabic character sequence to Latin CTS, \(T_4\) cannot reinterpret that sequence under a different language's rules. Consequently, a South Azerbaijani form embedded in a Turkish Wikipedia edition may be consumed by the first-pass Uyghur transliterator before the South Azerbaijani transliterator is reached. The same limitation applies to the Latin and Cyrillic proxy mappings. We regard this as a precision--coverage trade-off: the pipeline is conditioned on the primary corpus language but attempts to map as much mixed-script content as possible into a shared CTS-centered representation.

\subsection{Line Concatenation and Punctuation Augmentation}
\label{app:line-punctuation}

To avoid extremely short training contexts, we concatenated lines only within paragraph boundaries. Any line containing fewer than five tokens with at least one Unicode Letter character (general category \texttt{L}) was joined with subsequent lines until the combined sequence contained at least 20 such tokens or reached the end of the paragraph; all other lines were retained unchanged.

From each concatenated language file, we created two variants: one retained
all tokens, whereas the other discarded tokens containing neither a Unicode
Letter nor a Unicode Number character (general categories \texttt{L} and
\texttt{N}). The latter therefore removed standalone punctuation and symbol
tokens. This paired construction yielded 22 files for CTS and 22 for uroman:
11 line-concatenated files and 11 corresponding files with standalone
punctuation and symbol tokens removed. We then applied the same random seed
and identical shuffling-and-merging procedure separately to the corresponding
11-file sets in the CTS and uroman conditions.

\section{Character $n$-gram Coverage Metrics}
\label{app:ngram-coverage-metrics}

For a target token occurrence \(w_i\), let \(G_n(w_i)\) be the multiset of
all contiguous character $n$-gram occurrences extracted from
\texttt{<}\(w_i\)\texttt{>}; repeated $n$-grams at different positions are
retained. Let \(R_n\) be the set of character $n$-gram types occurring
anywhere in the relevant reference training tokens.

\paragraph{Token-mean coverage.}
Let \(N_n\) be the number of target token occurrences that yield at least one
$n$-gram. Token-mean coverage assigns equal weight to each such token:
\begin{equation}
C_{\mathrm{token}}(n)
=
\frac{1}{N_n}
\sum_{i:\lvert G_n(w_i)\rvert>0}
\frac{
\sum_{g\in G_n(w_i)}
\mathbf{1}[g\in R_n]
}{
\lvert G_n(w_i)\rvert
}.
\label{eq:ngram-token-mean-coverage}
\end{equation}

\paragraph{Occurrence coverage.}
Occurrence coverage assigns equal weight to every target $n$-gram position:
\begin{equation}
C_{\mathrm{occ}}(n)
=
\frac{
\sum_i\sum_{g\in G_n(w_i)}
\mathbf{1}[g\in R_n]
}{
\sum_i\lvert G_n(w_i)\rvert
}.
\label{eq:ngram-occurrence-coverage}
\end{equation}

\paragraph{Type coverage.}
Type coverage first deduplicates the target $n$-grams and then assigns equal
weight to every distinct type:
\begin{equation}
\begin{aligned}
T_n
&=
\bigcup_i
\operatorname{set}\left(G_n(w_i)\right),\\
C_{\mathrm{type}}(n)
&=
\frac{\lvert T_n\cap R_n\rvert}{\lvert T_n\rvert}.
\end{aligned}
\label{eq:ngram-type-coverage}
\end{equation}

For the pooled 2--5-gram values, matched and total occurrences or types are
combined across orders 2 through 5 before the corresponding coverage ratio is
computed. The pooled values are therefore not arithmetic means of the four
order-specific percentages.

\section{Apostrophe Normalization}
\label{app:apostrophe-normalization}

Table~\ref{tab:apostrophe-normalization} gives the complete
character-level mapping used during preprocessing. Every listed source
code point was replaced with the ASCII apostrophe (U+0027) before the
CTS and uroman processing branches diverged.

\begin{table}[t]
\centering
\small
\begin{tabular}{@{}l p{0.68\columnwidth}@{}}
\toprule
Source code point & Unicode character name \\
\midrule
\texttt{U+02BB} & MODIFIER LETTER TURNED COMMA \\
\texttt{U+2018} & LEFT SINGLE QUOTATION MARK \\
\texttt{U+2019} & RIGHT SINGLE QUOTATION MARK \\
\texttt{U+201B} & SINGLE HIGH-REVERSED-9 QUOTATION MARK \\
\texttt{U+0060} & GRAVE ACCENT \\
\texttt{U+00B4} & ACUTE ACCENT \\
\texttt{U+02BC} & MODIFIER LETTER APOSTROPHE \\
\texttt{U+FF07} & FULLWIDTH APOSTROPHE \\
\texttt{U+275B} & HEAVY SINGLE TURNED COMMA QUOTATION MARK ORNAMENT \\
\texttt{U+275C} & HEAVY SINGLE COMMA QUOTATION MARK ORNAMENT \\
\bottomrule
\end{tabular}
\caption{The ten source characters normalized to the ASCII apostrophe
(U+0027).}
\label{tab:apostrophe-normalization}
\end{table}

\section{fastText Training Configuration}
\label{app:fasttext-configuration}

The CTS and uroman fastText models were trained with
\texttt{fasttext.train\_unsupervised} on
\texttt{mergeshuffle\_cts\_file\_all} and
\texttt{mergeshuffle\_uroman\_file\_all}, respectively.
Table~\ref{tab:fasttext-configuration} reports every argument explicitly
passed to the training function. The two calls differed only in the input
corpus and the data-dependent bucket size. Arguments not explicitly supplied
to the function retained the defaults of the fastText library.

\begin{table}[t]
\centering
\small
\begin{tabular}{@{}lrr@{}}
\toprule
Argument & \texttt{cts\_ft} & \texttt{uroman\_ft} \\
\midrule
\texttt{model}    & \multicolumn{2}{c}{\texttt{cbow}} \\
\texttt{dim}      & 300 & 300 \\
\texttt{minCount} & 5 & 5 \\
\texttt{minn}     & 2 & 2 \\
\texttt{maxn}     & 5 & 5 \\
\texttt{bucket}   & 21,000,000 & 15,000,000 \\
\texttt{epoch}    & 10 & 10 \\
\texttt{lr}       & 0.05 & 0.05 \\
\texttt{ws}       & 5 & 5 \\
\texttt{neg}      & 10 & 10 \\
\texttt{loss}     & \multicolumn{2}{c}{\texttt{ns}} \\
\texttt{thread}   & 18 & 18 \\
\texttt{verbose}  & 2 & 2 \\
\bottomrule
\end{tabular}
\caption{Arguments explicitly supplied to
\texttt{fasttext.train\_unsupervised} when training the CTS and uroman
fastText models.}
\label{tab:fasttext-configuration}
\end{table}

\section{POS Dataset Split Statistics}
\label{app:pos-data-splits}

Table~\ref{tab:ud-pos-splits} reports the sentence counts for the 19 UD
v2.18 treebanks used in the POS experiments. We treated the UD
\texttt{dev} splits as validation data. Training and validation sentences
were pooled across treebanks, whereas evaluation was performed separately
on each available test split.

\begin{table}[t]
\centering
\small
\setlength{\tabcolsep}{4pt}
\begin{tabular}{@{}lrrr@{}}
\toprule
Treebank & Train & Validation & Test \\
\midrule
\texttt{az\_tuecl}       &      0 &     0 &   148 \\
\texttt{kk\_ktb}         &     31 &     0 & 1,047 \\
\texttt{ky\_ktmu}        &  1,308 &     0 & 1,222 \\
\texttt{ky\_tuecl}       &      0 &     0 &   173 \\
\texttt{tr\_atis}        &  4,126 &   572 &   586 \\
\texttt{tr\_boun}        &  7,803 &   979 &   979 \\
\texttt{tr\_framenet}    &  2,288 &   205 &   205 \\
\texttt{tr\_gb}          &      0 &     0 & 2,880 \\
\texttt{tr\_imst}        &  3,435 & 1,100 & 1,100 \\
\texttt{tr\_kenet}       & 15,398 & 1,646 & 1,643 \\
\texttt{tr\_penn}        & 14,849 &   622 &   924 \\
\texttt{tr\_pud}         &      0 &     0 & 1,000 \\
\texttt{tr\_tourism}     & 15,473 & 2,166 & 2,191 \\
\texttt{tr\_tuecl}       &      0 &     0 &   148 \\
\texttt{tt\_nmctt}       &      0 &     0 &   148 \\
\texttt{ug\_udt}         &  1,656 &   900 &   900 \\
\texttt{uz\_tuecl}       &      0 &     0 &   148 \\
\texttt{uz\_ut}          &      0 &     0 &   500 \\
\texttt{uz\_uzudt}       &    483 &     0 &   201 \\
\midrule
Pooled total             & 66,850 & 8,190 & 16,143 \\
\bottomrule
\end{tabular}
\caption{Sentence counts by split for the 19 UD v2.18 treebanks used in
the POS experiments. UD \texttt{dev} splits are listed under
\emph{Validation}.}
\label{tab:ud-pos-splits}
\end{table}

\section{Significance Tests against CANINE-c}
\label{app:canine-significance}

Table~\ref{tab:canine-significance} reports the treebanks on which a
fastText-based system achieves a higher mean than CANINE-c with an
uncorrected $p<0.05$.

\begin{table*}[t]
\centering
\scriptsize
\setlength{\tabcolsep}{4pt}
\renewcommand{\arraystretch}{0.96}
\begin{tabular}{@{}lllccrrr@{}}
\toprule
Metric & Transliteration & Treebank &
fastText score & CANINE-c & $\Delta$ &
Raw $p$ & Holm $p$ \\
\midrule
Accuracy & CTS &
\texttt{tr\_boun} &
$90.36 \pm 0.32$ & $89.80 \pm 0.28$ &
$+0.56$ & 0.0215 & 0.1288 \\

Accuracy & CTS &
\texttt{tr\_gb} &
$89.54 \pm 0.24$ & $88.53 \pm 0.06$ &
$+1.01$ & 0.0003 & $\mathbf{0.0052}$ \\

Accuracy & CTS &
\texttt{tr\_pud} &
$83.64 \pm 0.18$ & $82.90 \pm 0.20$ &
$+0.75$ & 0.0004 & $\mathbf{0.0071}$ \\

Accuracy & CTS &
\texttt{ug\_udt} &
$87.28 \pm 0.58$ & $86.23 \pm 0.27$ &
$+1.05$ & 0.0162 & 0.1136 \\

Accuracy & CTS &
\texttt{uz\_tuecl} &
$76.85 \pm 0.95$ & $74.82 \pm 0.50$ &
$+2.02$ & 0.0250 & 0.1288 \\

Accuracy & CTS &
\texttt{uz\_uzudt} &
$86.85 \pm 0.70$ & $84.43 \pm 0.34$ &
$+2.42$ & 0.0025 & $\mathbf{0.0352}$ \\
\midrule
Accuracy & uroman &
\texttt{tr\_boun} &
$90.52 \pm 0.23$ & $89.80 \pm 0.28$ &
$+0.72$ & 0.0057 & 0.0688 \\

Accuracy & uroman &
\texttt{tr\_gb} &
$89.47 \pm 0.26$ & $88.53 \pm 0.06$ &
$+0.94$ & 0.0021 & $\mathbf{0.0299}$ \\

Accuracy & uroman &
\texttt{tr\_pud} &
$83.71 \pm 0.08$ & $82.90 \pm 0.20$ &
$+0.81$ & 0.0010 & $\mathbf{0.0149}$ \\

Accuracy & uroman &
\texttt{uz\_uzudt} &
$85.92 \pm 0.39$ & $84.43 \pm 0.34$ &
$+1.49$ & 0.0001 & $\mathbf{0.0027}$ \\
\midrule
Macro-$F_1$ & CTS &
\texttt{tr\_tuecl} &
$77.31 \pm 0.58$ & $75.89 \pm 0.77$ &
$+1.43$ & 0.0143 & 0.1855 \\

Macro-$F_1$ & CTS &
\texttt{ug\_udt} &
$76.94 \pm 1.23$ & $70.67 \pm 0.81$ &
$+6.27$ & 0.0011 & $\mathbf{0.0165}$ \\

Macro-$F_1$ & uroman &
\texttt{ug\_udt} &
$73.22 \pm 1.02$ & $70.67 \pm 0.81$ &
$+2.56$ & 0.0162 & 0.2230 \\
\bottomrule
\end{tabular}
\caption{FastText-based POS results that exceed CANINE-c with an
uncorrected $p<0.05$. Scores are means $\pm$ sample standard deviations
over five seed-aligned runs. $\Delta$ is the fastText-based score minus
the CANINE-c score. Raw $p$ values come from two-sided paired $t$-tests.
Holm correction is applied across all 19 treebanks separately for each
representation--metric family, including comparisons not displayed in
this table. Bold Holm values indicate corrected $p<0.05$.}
\label{tab:canine-significance}
\end{table*}

\section{Artifact Licensing}
\label{app:artifact-licensing}

The released transliterated Wikipedia corpora are distributed in
accordance with the applicable Wikimedia Creative Commons
Attribution--ShareAlike terms, with attribution, license notices, and
the transliteration modifications documented. We do not redistribute
WikiANN or Universal Dependencies data; users must obtain these datasets
from their original sources and comply with their respective licenses.
The released code, configurations, and model artifacts include their
applicable licenses and third-party notices.

\section{Use of AI Assistance}
\label{app:ai-assistance}
AI-assisted tools were used during manuscript preparation for language editing, improving clarity and concision, and limited assistance with \LaTeX{} formatting. They were not used to generate the research idea, formulate hypotheses, design or conduct experiments, or make scientific interpretations and conclusions. All AI-assisted revisions were reviewed and verified by the authors, who take full responsibility for the content of this work.

\end{document}